\documentclass[a4paper,twocolumn]{article}

\usepackage[english]{babel}
\usepackage[utf8x]{inputenc}
\usepackage[T1]{fontenc}

\usepackage{amsmath}
\usepackage{amsfonts}
\usepackage{amssymb}
\usepackage{amsthm}

\usepackage{graphicx} 
\usepackage{bmpsize}
\usepackage{hyperref}
\usepackage{url} 

\usepackage{tikz}

\theoremstyle{definition}

\usepackage{booktabs} 
\usepackage{multirow}

\usepackage{caption}
\usepackage{subcaption}

\usepackage{listings}
\usepackage[figure]{algorithm2e}

\title{The Helmholtz Method:\\ Using Perceptual Compression to Reduce Machine Learning Complexity}

\author{Gerald Friedland\footnote{also at Lawrence Livermore National Laboratory}, Jingkang Wang, Ruoxi Jia, Bo Li\\University of California, Berkeley\\
fractor@eecs.berkeley.edu, wangjksjtu\_01@sjtu.edu.cn, \\ruoxijia@berkeley.edu, lxbosky@gmail.com
} 

\date{July 5th, 2018}

\begin{document}

\maketitle

%
%

\begin{abstract}
This paper proposes a fundamental answer to a frequently asked question in multimedia computing and machine learning: Do artifacts from perceptual compression contribute to error in the machine learning process and if so, how much? Our approach to the problem is a reinterpretation of the Helmholtz Free Energy formula from physics to explain the relationship between content and noise when using sensors (such as cameras or microphones) to capture multimedia data. The reinterpretation allows a bit-measurement of the noise contained in images, audio, and video by combining a classifier with perceptual compression, such as JPEG or MP3. Our experiments on CIFAR-10 as well as Fraunhofer's IDMT-SMT-Audio-Effects dataset indicate that, at the right quality level, perceptual compression is actually not harmful but contributes to a significant reduction of complexity of the machine learning process. That is, our noise quantification method can be used to speed up the training of deep learning classifiers significantly while maintaining, or sometimes even improving, overall classification accuracy. Moreover, our results provide insights into the reasons for the success of deep learning.
\end{abstract}

\section{Introduction}
As datasets for multimedia evaluations~\cite{larson2012community,awad2017trecvid} grow larger and larger and become more difficult to handle, performing machine learning on perceptually compressed data has become increasingly mainstream~\cite{ni2015large}. In the past practice of feature extraction, many communities had established the rule, however, that features should be extracted on uncompressed data for machine learning to achieve better results. Today, with the exception of consumer produced videos, machine learning systems are usually trained and tested on high quality image and audio datasets. So input quality is an important practical challenge that is often overlooked in the design of machine learning systems. Deep learning has obtained state-of-the-art performance on many machine vision and audition tasks. However, the relationship between lossy compression and machine learning performance mostly remains unexplored.  In this article, we propose a fundamental answer to this question. Our approach to the problem is a reinterpretation of the Helmholtz free energy formula from physics to explain the relationship between content and noise when using sensors to capture multimedia data. This allows the empirical estimation of the noise content in images, audio, and videos based on combining a classifier with perceptual compression, such as JPEG or MP3. Our experiments on the CIFAR-10 as well as Fraunhofer's IDMT-SMT-Audio-Effects dataset indicate that, at the right quality level, perceptual compression is not harmful and contributes to a significant reduction of complexity of the machine learning process. That is, our noise estimation method can be used to speed up the training of deep learning classifiers significantly while maintaining and sometimes even improving overall classification accuracy. Moreover, we conclude that our results provide insights into the reasons for the success of deep learning.

\begin{figure}
    \centering
        \includegraphics[width=0.5\textwidth]{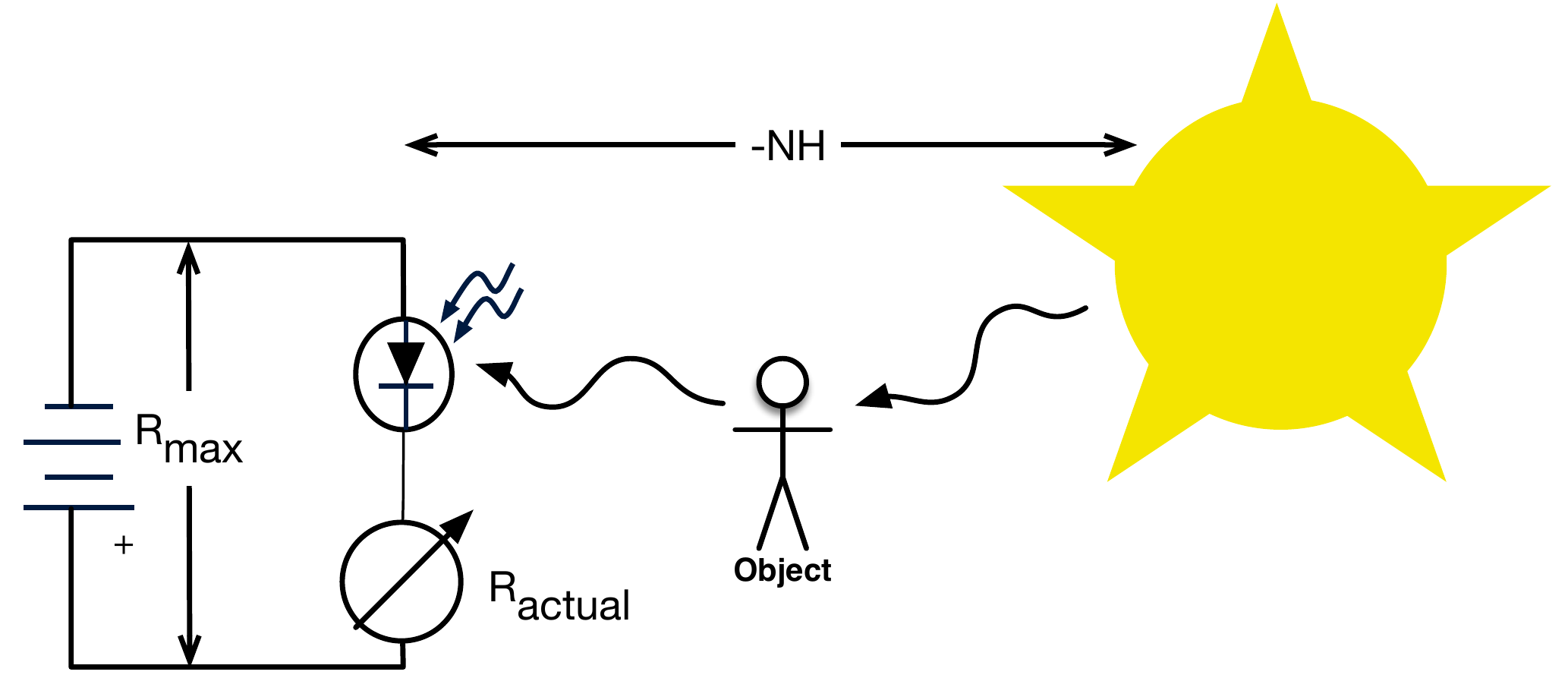}
\caption{\label{fig:helmholtz}Our interpretation of the Helmholtz free energy equation $A \equiv U-TS$ applied to camera sensors. Potential energy of the battery ($U$) is converted into kinetic energy $A$ (electron flow) in response to light hitting the sensor. The solar energy, however, is reduced by hitting objects it reflects from, changing both frequency (chrominance) and amplitude (luminance) of the light wave. $A$ denotes to the energy variation in the circuit. Assuming a constant sampling rate, the measured pixel $R_{actual}$ corresponds to $A$ divided by the sampling period length. $R_{max}$ corresponds to $U$ in the same way. We reinterpret $-TS$ as $-NH$. This is, any noise $N$ increases the minimum description length $H$ of the pixel (energy reduction of the light).
}
\end{figure}

Our paper is structured as follows: Section~\ref{sec:related} starts by presenting related work on the interaction between perceptual compression and machine learning and also on the use of Helmholtz' theories to to inspire results in machine learning and multimedia applications. Section~\ref{sec:helmholtz} briefly introduces the original physical interpretation of Helmholtz free energy and then derives our re-interpretation of the formula. Section~\ref{sec:perceptual} explains how the re-interpretation can be used productively to estimate noise content, culminating in the derivation of the expected shape of the measurement results in Figure~\ref{fig:qopttheory}. Section~\ref{sec:experiments} then presents measurements on both audio and image data to confirm the theoretical derivations. Section~\ref{sec:conclusion} finally concludes the article with an interpretation of the nature of deep learning in the light of the results in this paper and presents possible directions for future work.

\section{Related Work}
\label{sec:related}
Compression techniques are widely used in machine learning for improving scalability, generalizability, and robustness of models. In the past years, various feature selection algorithms have been developed to tease out crucial features from high-dimensional data, such as images, DNA sequences, and audio signals. It has been demonstrated both theoretically and empirically that removing noisy and redundant features can not only lower computational complexity, but improve learning performance~\cite{peleg2005feature}. Autoencoders are another popular type of compression method that directly learn a low-dimensional feature space from data~\cite{kingma2013auto}. It is worth noting that Hinton et al.~\cite{hinton1994autoencoders} presented an objective function for training Autoencoders, which has exactly the form of Helmholtz free energy. Compression can also be implicitly embedded in the training process. For instance, sparse learning such as $l_1$ regularization~\cite{nishimoto2011reconstructing} enforces model coefficients to be small or exactly zero; as a result, only very few feature dimensions are active to make predictions. It has been shown recently in~\cite{shwartz2017opening} that standard training of deep neural networks embodies a compression phase where hidden layers gradually output finds a succinct representation of input features and reduces irrelevant information until convergence.

The interaction of perceptual compression and machine learning is not very well studied, however. Dodge et al.~\cite{dodge2016understanding} demonstrates that the performance of deep neural networks is ``surprisingly" robust to artifacts introduced by perceptual compression, compared with other types of image distortions such as blurring and random noise. As a note, perceptual compression has also proved useful for improving models' robustness against adversarial example-signals that are intentionally made close to natural multimedia signals but misclassified by models~\cite{das2017keeping}.

In this article, we present a theoretical model that describes the interaction of sensor data with machine learning along with empirical evidence to verify the theoretical hypotheses. Furthermore, we show how, using the methodology employed in this paper, it is possible to estimate the amount of noise versus the amount of content in sensor data.

\section{Helmholtz Free Energy}
\label{sec:helmholtz}

In thermodynamics, Helmholtz free energy is a thermodynamic potential that measures the ``useful" work obtainable from a closed thermodynamic system at a constant temperature and volume. The Helmholtz free energy is defined as
\begin{equation}
 A\equiv U-TS,
\end{equation}
where $A$ is the Helmholtz free energy, $U$ is the internal energy of the system, $T$ is the absolute temperature of the surroundings, and $S$ is the physical entropy of the system. Since it's original publication in 1882~\cite{Helmholtz1888} it has been found many times that this formula is generally useful to describe the influence of an externality on a thermodynamically closed system (this is, a system that exchanges energy but not matter with its surroundings). Most importantly, the Helmholtz formula is often used to define fundamental equations of state and the key variables are interpreted as probability distributions. Inspired by this notion, we reinterpret the formula for the purposes of sensor measurements as they frequently occur in the field of Multimedia Computing and show that this re-interpretation is useful. 

A pixel is usually treated as a signal which is a physically speaking a power. Energy is power multiplied by time. Treating measurements as signals is convenient for computation but in reality any sensor requires activation energy, this is, a signal through time. To resolve this impreciseness, for this deduction, we assume that frame rates, capture times, and/or sampling rates are constant. This is usually true for multimedia capturing devices such as photo and video cameras, or digital microphones. This assumption is convenient as it makes the signal treatable as an energy divided by a constant. 

We re-purpose the Helmholtz formula as follows:
\begin{equation}
\label{eq:newhelmholtz}
R_{actual}=R_{max}-NH,
\end{equation}
where $R_{actual}$ is the actual sensor reading (for example, actual pixel value), $R_{max}$ is the maximum sensor reading (for example, the reference white of an image sensor), $N$ is an unknown scalar that characterizes the noise effects in the measurement process (for example, for image sensors this includes everything from camera lens aberration to dead pixels but also actual temperature variation), and $H$ is the information captured by the sensing process.  Consistent with the original formula, we treat $R_{actual}$ and $R_{max}$ as energies and $H$ is the Shannon entropy~\cite{shannon1948mathematical}. As explained, all energies can be converted to signals by dividing by the constant sampling period. For convenience, we assume this constant to be $1$. 

Figure~\ref{fig:helmholtz} exemplifies the idea using an image sensor. A measured pixel ($R_{actual}$) is in reality the energy collected by a camera sensor over a certain time (shutter time) $A$. The reference white $R_{max}$ is the maximum value that can be measured (corresponding to internal energy $U$). That is, we assume that if the photo element was directly pointed at the sun, then $R_{actual}=R_{max}$. However, cameras usually capture the reflection of a sun ray; as a result, the photo element is influenced by an externality. As the sunlight is reflected from a surface of a certain chemical composition, it loses energy and therefore reduces $R_{max}$ by a certain value. According to Helmholtz, this value is the factor of two components: 1) An entropy term $S$ (unknown a-priori to the photo sensor circuit), which we call $H$ in order to follow the naming convention of the Shannon entropy and 2) temperature $T$, which we generalize as the noise constant $N$. 


In the case of audio, $R_{max}$ captures the maximum energy that the microphone can record to generate one sample. $H$ is the information contained in the sample and $N$ can be treated as noise.
Similar analogies can be made for other sensors. In the following sections we use this model to explain why perceptual compression works and how this model can be exploited to safely reduce the number of parameters in any machine learning model. 

\section{Perceptual Compression}
\label{sec:perceptual}
\begin{figure}
\includegraphics[angle=270,width=0.48\textwidth]{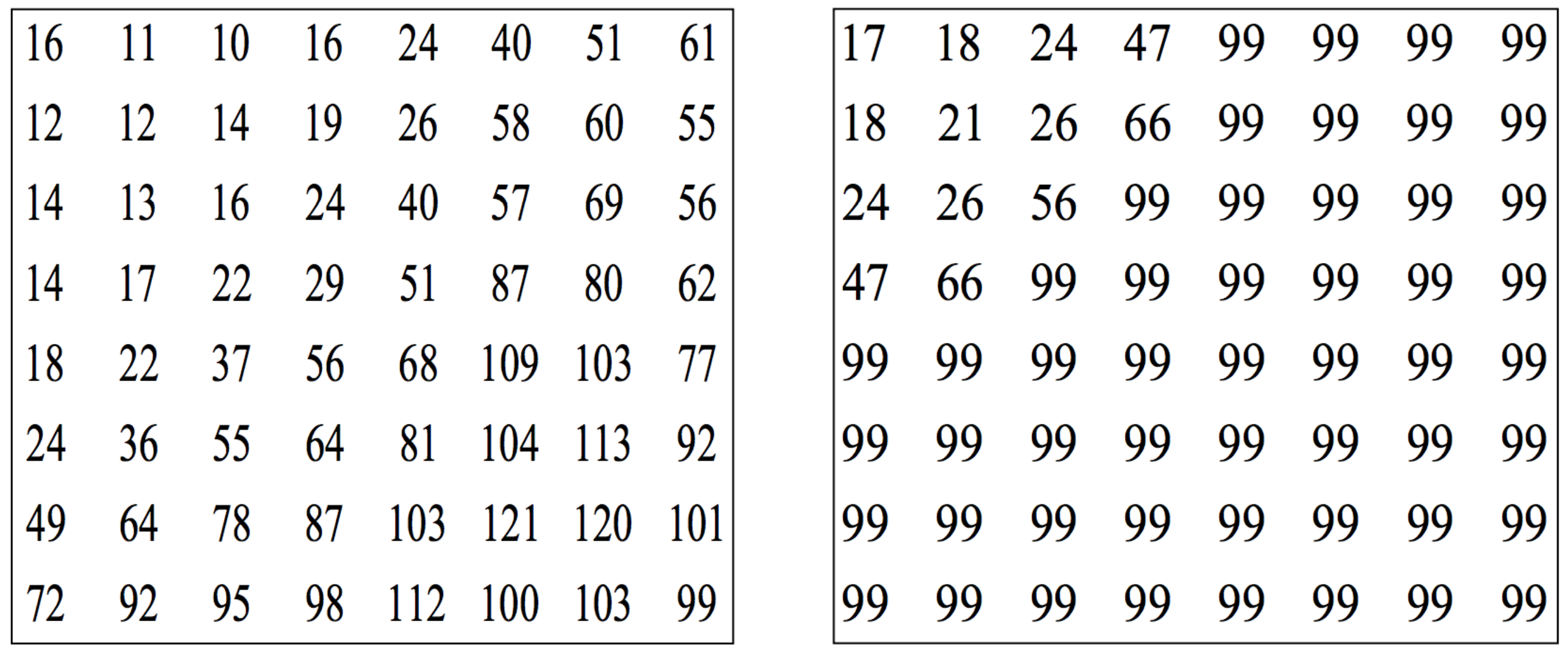}
\caption{\label{fig:quantization} Quantization matrices K.1 (left, luminance component) and K.2 (right, chrominance components) as defined by the JPEG standard~\protect\cite{jpeg1992}. This corresponds to an average loss of 13.6 bits per pixel at JPEG quality level 50.}
\end{figure}
Most, if not all, images used for typical benchmark sets, such as ImageNet~\cite{deng2009imagenet} or YFCC100M~\cite{thomee2016yfcc100m}, are compressible using JPEG, in fact, most of them come as JPEGs. JPEG is the popular name for the JFIF format, a commonly used method of lossy compression of digital images, particularly photographs. The degree of compression can be adjusted, allowing to control the tradeoff between storage size and image quality. JPEG typically achieves 10:1 compression with little perceptible loss in image quality. We now take a closer look at the underpinnings of JPEG and how they relate to the Helmholtz free energy formula.

In a nutshell~\cite{jpeg1992}, JPEG incurs several losses. First, a discrete cosine transform (DCT) converts each of the three $8\times8$ blocks per channel (Y, U, V) from the spatial domain into the frequency domain. By default, this transformation is reversible but JPEG applies rounding so several low-significance bits may be lost. For the purposes of this article, we ignore this loss. Then, a perceptual model based loosely on the human psychovisual system discards high-frequency information using quantization on the DCT coefficients. The quantized coefficients are then serialized and entropy encoding is applied to produce the final output. Decompression unpacks the bits, multiplies with the same quantization matrix and reverses the DCT before converting the pixels from (Y, U, V) into (R, G, B) again for display.

This transformation fits our Helmholtz model as follows. As explained in Section~\ref{sec:helmholtz}, our interpretation is that each pixel value $R_{actual}$ in each Y, U, or V dimension is represented as described in Equation~\ref{eq:newhelmholtz}. Since $R_{max}$ is a constant (usually $8$ bit, this is the signal corresponds to a pixel value of $255$), we disregard it and the sign and observe that the a pixel is composed of information content $H$ and a noise factor $N$. 

\begin{figure}
\includegraphics[width=0.48\textwidth]{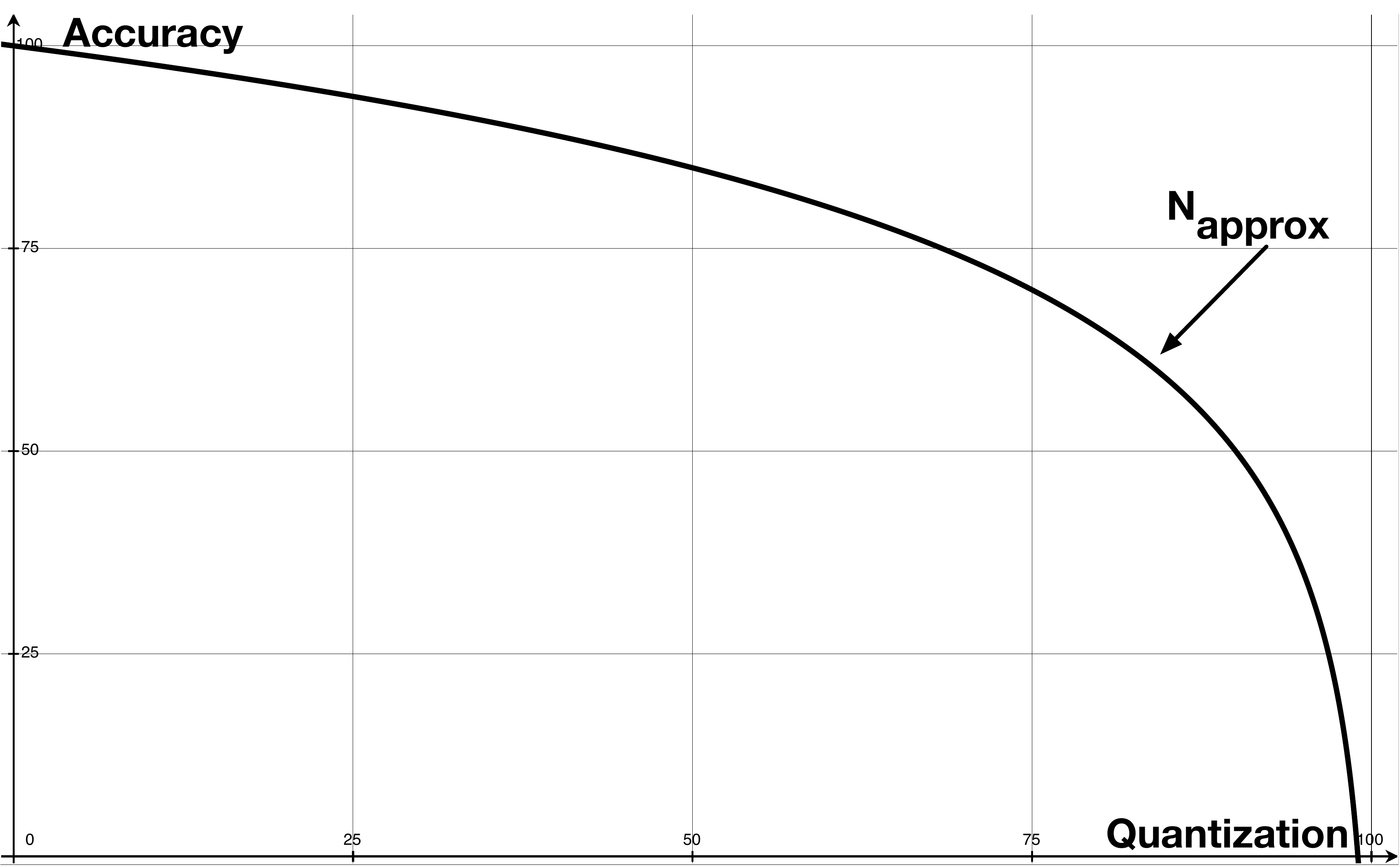}
\caption{\label{fig:qopttheory} Hypothesized accuracy/quantization curve as per the model derived from the Helmholtz Free Energy equation~\protect\ref{eq:newhelmholtz}. The Y-axis shows a hypothesized accuracy and the X-axis shows an increasing quantization factor. The sharp drop should occur at the point where the quantization approximates the noise most exactly. Note: The theory only predicts the slope and the axis scaling is arbitrarily chosen.}
\end{figure}

We can therefore obtain an estimate of the noise factor via the following equation:
\begin{equation}
\frac{NH}{N_{approx}}=H_{approx},
\end{equation}
where $N_{approx}$ is the approximated noise. $H_{approx}$ is then the approximated content. Just like in JPEG, $H_{approx}$ contains mostly clean information and can therefore be compressed using entropy compression. Most importantly for this article, as indicated by related work (see Section~\ref{sec:related}), the hypothesis is that $H_{approx}$ can be better modelled using any encoder. That is, it should be easier for a machine learner to model the mutual information between the input data $H_{approx}$ and some labels compared to modeling the mutual information between the direct input $NH$ and the same labels. As a consequence, we would expect that the resulting accuracy/compression curve looks like Figure~\ref{fig:qopttheory}. The point where the curve falls drastically would be $N_{approx}$. The curve is characterized by the fact that, if $N_{approx} > N$, the quantization will, intuitively speaking, ``cut into the content" and destroy information contained in $H$.

\subsection{Practical Implementation}
In the following, we detail the JPEG quantization process and calculate the loss incurred in bits so we can relate it to the experimental results. By default, the three RGB channels have each $8$ bit resolution, this is $R+G+B=24$bits. Converting to YUV space subsamples the chroma components $2:1$. That is, $24$ bits become $16$ bits. As explained earlier, we assume the discrete cosine transformation to be lossless. Therefore $8$ bit Y component and $4+4$ bits UV components are only reduced by quantization. The quantization of the Y component is performed by dividing the DCT coefficients by matrix K.1 and the quantization of the U and V components are performed by dividing by matrix K.2 (see Figure~\ref{fig:quantization}). The loss in bits can be easily calculated as $\log_2 12487=13.6$ bits, where $12487$ is the sum of all components of K.1 and K.2, with the components in K.2 counted twice. Even at the default quality level $q=80$ (libJPEG), each pixel is reduced by $12.2$ bits to a little less than $4$ bits per pixel. For scenarios with greater loss, for example quality level $q=25$ (see Section~\ref{sec:experiments}), the loss is even steeper. At this quality level, only $1.4$ bits per pixel remain. Arithmetically, we reach the level of $1$ bit average per pixel at quality level $19$. 

\begin{figure}
\includegraphics[width=0.48\textwidth]{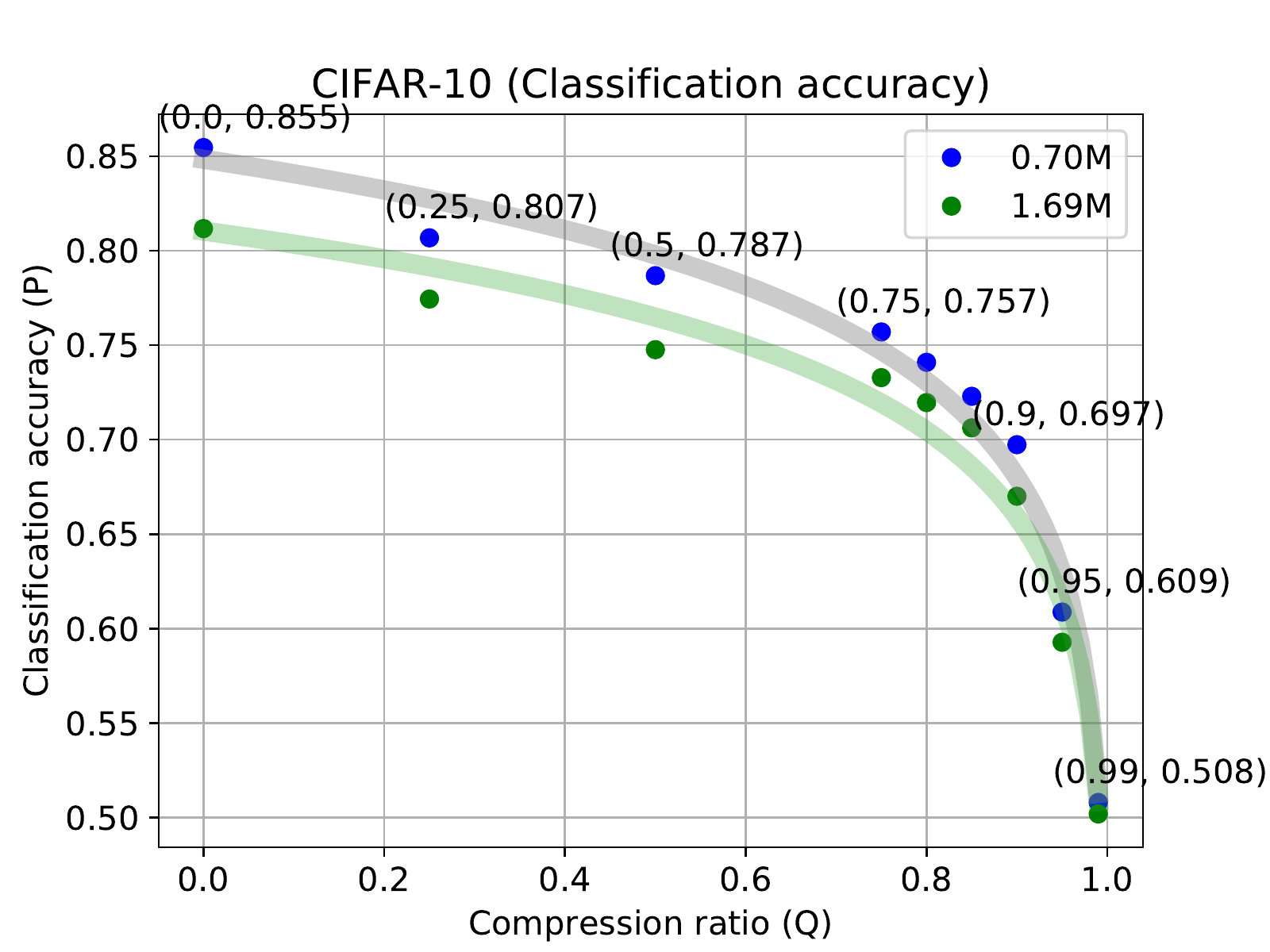}
\caption{\label{fig:cifar_acc} Classification accuracy on CIFAR-10 as a function of the JPEG compression quality $q$ based on the results of classifier setup A (blue dots) and classifier setup F (green dots). The shadow curves represent the properly scaled version of the theoretical curve from Figure~\protect\ref{fig:qopttheory}. For comparability with the audio results, the x-axis is scaled as $Q=\frac{100-q}{100}$  ($q$ being the JPEG quality level).}
\end{figure}

\begin{figure}
\includegraphics[width=0.48\textwidth]{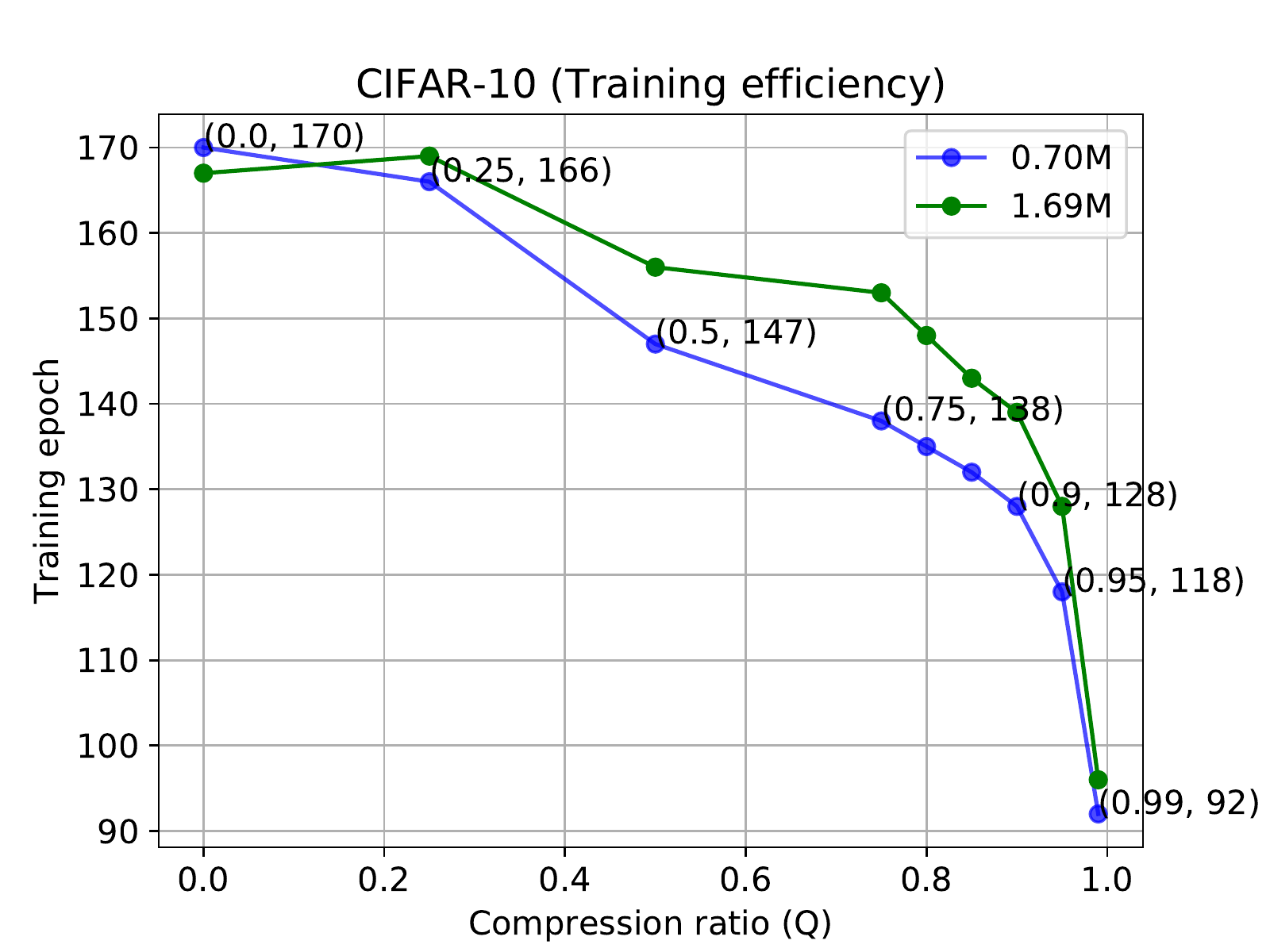}
\caption{\label{fig:cifar_time} Comparison of the number of required training epochs for the CIFAR-10 classification task as a function of the JPEG quality $q$. The x-axis is again scaled as $Q=\frac{100-q}{100}$.}
\end{figure}

\begin{table*}
	\centering
	\caption{Neural network architectures used on the CIFAR-10 classification task. Conv: convolutional layer, FC: fully connected layer, Global\_avg\_pooling: global average pooling layer.}
	\begin{tabular}{@{}ccc@{}}
    \toprule
		\textbf{A}  & \textbf{C} & \textbf{F}  \\ \hline \hline
		Conv({[32, 64]}, 3, 3) + ReLU   & Conv({[32, 64]}, 3, 3) + ReLU   & Conv({[32, 64]}, 3, 3) + ReLU   \\
		Conv(128, 3, 3) + Dropout(0.5)  & Conv(128, 3, 3) + Dropout(0.5)  & Conv(128, 3, 3) + Dropout(0.5)  \\
		Conv({[128, 128]}, 3, 3) + ReLU & Conv({[128, 128]}, 3, 3) + ReLU & Conv({[128, 128]}, 3, 3) + ReLU \\
		Conv(128, 3, 3) + Dropout(0.5)  & Conv(128, 3, 3) + Dropout(0.5)  & Conv(128, 3, 3) + Dropout(0.5)  \\
		Conv({[128, 128]}, 3, 3) + ReLU & Conv({[128, 128]}, 3, 3) + ReLU & Flatten                         \\
		Conv(10, 3, 3)                  & Conv(128, 3, 3) + Dropout(0.5)  & FC(128) + Dropout(0.5)          \\
		Global\_avg\_pooling            & Conv({[128, 128]}, 3, 3) + ReLU & FC(256) + Dropout(0.5)  \\
		Softmax                         & Conv(10, 3, 3)  				  & FC(256) + Dropout(0.5)   \\ 
		                                & Global\_avg\_pooling            & FC(10)                   \\
		 & Softmax & Softmax \\  \midrule
		701,386 (0.70M) & 1,144,138 (1.14M) & 1,686,090 (1.69M) \\ \bottomrule 
	\end{tabular}
	\label{tab:cifar_arch}
\end{table*}

From these calculations, we can conclude that most of the visual content of an image is actually stored in the highest significant bits and the lower significant bits only contain noise. In order to get an estimate of the content portion relevant for classification, we therefore train a classifier with identical input images, except different quality levels of JPEG. We then expect the typical training error of the classification to drop steeply at the point where the quantization gets too high. That is, at the point where the chosen quality level $q$ implies $N_{approx} > N$. Choosing an optimal $q$, however, should lead to overall less training time because the number of parameters for the machine learner can be reduced as the complexity induced by noise does not have to be modeled. Figure~\ref{fig:qopttheory} illustrates the idea. The curve shown is generated as $Accuracy=c*\log(100-Q)$ where $c$ is a scaling constant and $Q$ is quantization in percent. $\log(100-Q)$ stands for the information content of a compressed image. From Fano's inequality~\cite{cover2012elements}, the upper bound on the classification accuracy is linear in the entropy of the image.

Acoustic perceptual compression works conceptually similar to visual compression~\cite{friedland2014}. However, the ear is more sensitive to distortions than the eye. Therefore, a more accurate version of the DCT, the mDCT~\cite{le1991mpeg}, is usually applied and the quantization matrices are not linear but tuned to the human auditory system using empirical measurements. That is, our theory remains unchanged for acoustic signals but we expect our hypothesized curve to be less accurate due to the non-linear quantization (compare also Figure~\ref{fig:audio_acc}).

\section{Experimental Results}
\label{sec:experiments}
In the following, we provide empirical evidence to confirm the theoretical derivations from the previous sections. We conduct extensive evaluations on the interaction of machine learning models and perceptual compression for both image and audio classification tasks. Our experiments have two goals. First, we want to confirm the shape of the curve predicted in Figure~\ref{fig:qopttheory} and, as a result, being able to confirm the possibility of measuring $N_{approx}$. Second, we wanted to see if the knowledge of $N_{approx}$ contributes to a reduction of complexity of machine learning models.

\subsection{Datasets}
For our image experiments, we use the CIFAR-10 image dataset~\cite{krizhevsky2009learning}, which consists of $60000$ $32 \times 32$ color images in $10$ classes, with $6000$ images per class. There are $50000$ training images and $10000$ test instances. We then apply different levels of JPEG compression to correlate the compression ratio with the classification accuracy for the CIFAR-10 task.
For the audio experiments, we leverage the Fraunhofer IDMT-SMT-Audio-Effects dataset~\cite{stein2010automatic}, which is a large database for automatic detection of audio effects in recordings of electric guitar and bass and related signal processing \footnote{\url{https://www.idmt.fraunhofer.de/en/business_units/m2d/smt/audio\_effects.html}}. This dataset contains $55044$ uncompressed WAV files ($44.1$\,kHz, $16$\,bit, mono) with single recorded notes. In our experiments, we explore a subset of $12$ classes of different audio effects and $20592$ monophonic guitar samples in total, with $75$\% for training and $25$\% for testing.

Note that, due to space constraints, this article only presents a subset of all our experiments. Therefore, as outlined in Section~\ref{sec:conclusion}, the full set of experiments is available online for reproducibility.

\subsection{Image Classification}
\textbf{Preprocessing}: To make our results comparable, we follow previous works~\cite{SpringenbergDBR14, MishkinM15} and normalize the images and adopt the same data augmentation including random shifting and flipping. It is worth noting that the input of our neural networks follows the  original $32 \times 32$ image definition provided by CIFAR-10. That is, we do not resize images for our experiments. 

\noindent\textbf{Model Architectures}: Our experiments are designed to reduce complexity of networks while maintaining high performance. Overall, we explore six model architectures (A to F) with different number of parameters (0.7M to 1.69M), covering several classic architectures of deep neural networks. Table~\ref{tab:cifar_arch} shows three representative settings. The models are able to handle data of different complexity based on the number of parameters used. They are: 

\begin{itemize}
\item A: Architecture similar to All Convolutional Net~\cite{SpringenbergDBR14}, where no fully connected layers are employed, but replace first three convolutional layers with VGG~\cite{SimonyanZ14a} setting (channels 32, 64, 128). Size of parameters: 0.70M. (\textbf{Blue dots in Figure~\ref{fig:cifar_acc} and ~\ref{fig:cifar_time}})
\item B: Multiple fully connected layers following convolutional layers are added based on A. Size of parameters: 1.08M
\item C: Extra convolutional group (three 128 channel convolutional layers) is extended based on A. Size of parameters: 1.14M 
\item D: Both convolutional group and multiple fully connected layers are added based on A. Size of parameters 1.28M
\item E: Multiple fully connected layers with larger units are adopted compared to B. Size of parameters: 1.62M
\item F: More fully connected layers are extended compared to E. Size of parameters: 1.69M (\textbf{Green dots in Figure~\ref{fig:cifar_acc} and ~\ref{fig:cifar_time}})
\end{itemize}

\begin{figure}
\includegraphics[width=0.48\textwidth]{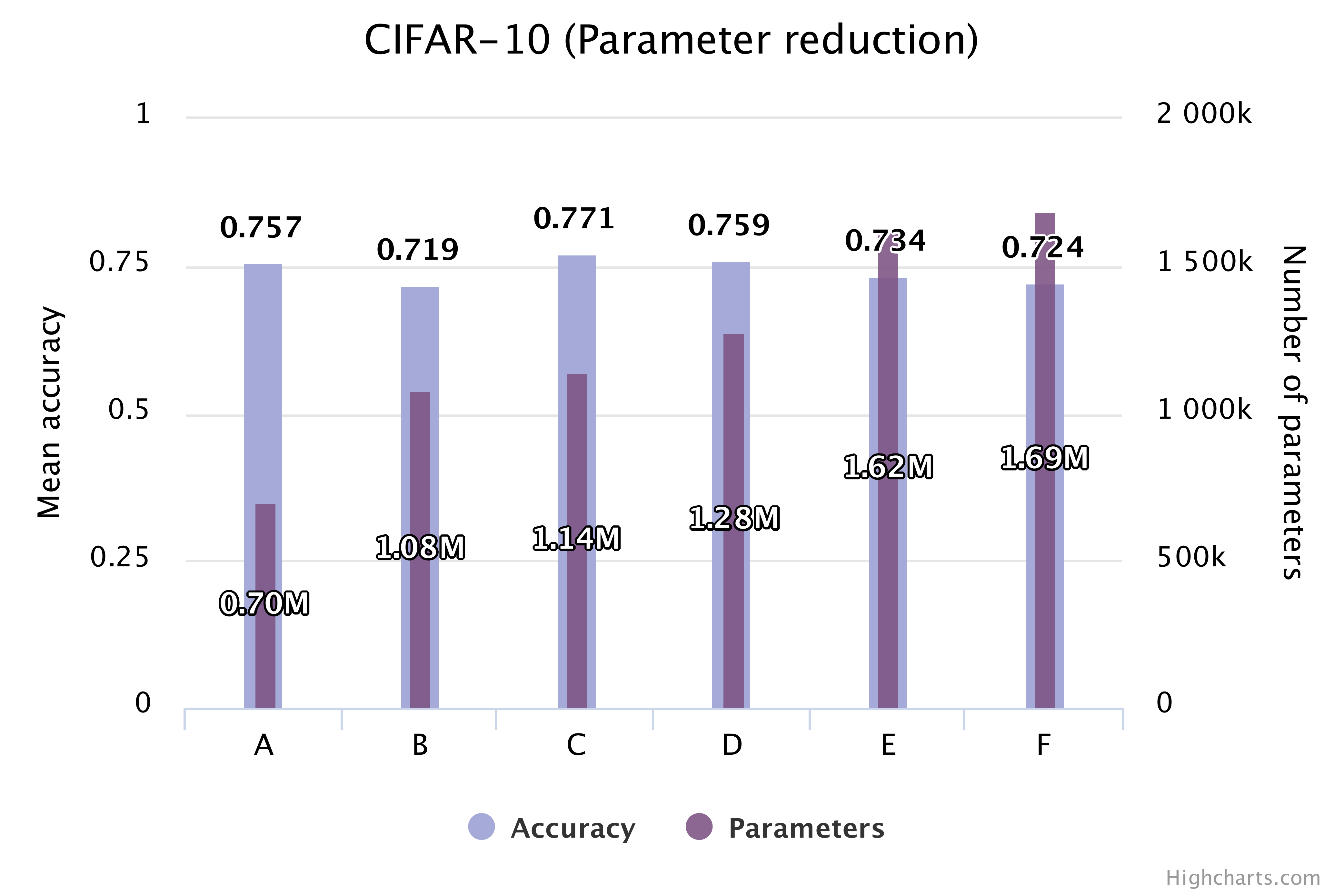}
\caption{\label{fig:cifar_paras} Classification accuracy on CIFAR-10 as a function of parameter size using aggressively compressed JPEG images (quality level $q=25$).}
\end{figure}

Figure~\ref{fig:cifar_paras} evaluates different effects of parameter reduction methods, and based on the comparison, we use A to demonstrate the compression effects under different ratios. A similar trend can be observed for different parameter reduction methods.

\subsection{Audio Classification}

\textbf{Preprocessing}: We perform compression on the input WAV files using the open-source MPEG-Audio Layer 3 implementation LAME \footnote{\url{http://lame.sourceforge.net/}}. In contrast to JPEG, LAME is not parameterized using a quality level $q$ but using a target bitrate. We then generate mel-spectrograms of all audio files and the number of mel coefficients used in spectrograms is set to $96$. 

\noindent\textbf{Architectures}: Similar to the image experiments, we explore six model architectures (A to F) with different numbers of parameters (0.10M to 3.72M) for the audio classification task. Table~\ref{tab:audio_arch} shows three representative settings. Our performance is comparable to the baseline reported in \cite{stein2010automatic} where a Support Vector Machine is trained. We define two loops: 1) convolutional layer loop (Conv-loop) that is composed of a convolutional layer, an ELU \cite{ClevertUH15} nonlinear activation layer, a max-pooling layer and a dropout layer; 2) fully connected layer loop (FC-loop) that is composed of a fully connected layer with $128$ units and a dropout regularization layer (compare Table~\ref{tab:audio_arch}). For better performance, the dropout rates are set to $0.5$ and $0.6$ in two loops, respectively.

\begin{figure}
\includegraphics[width=0.48\textwidth]{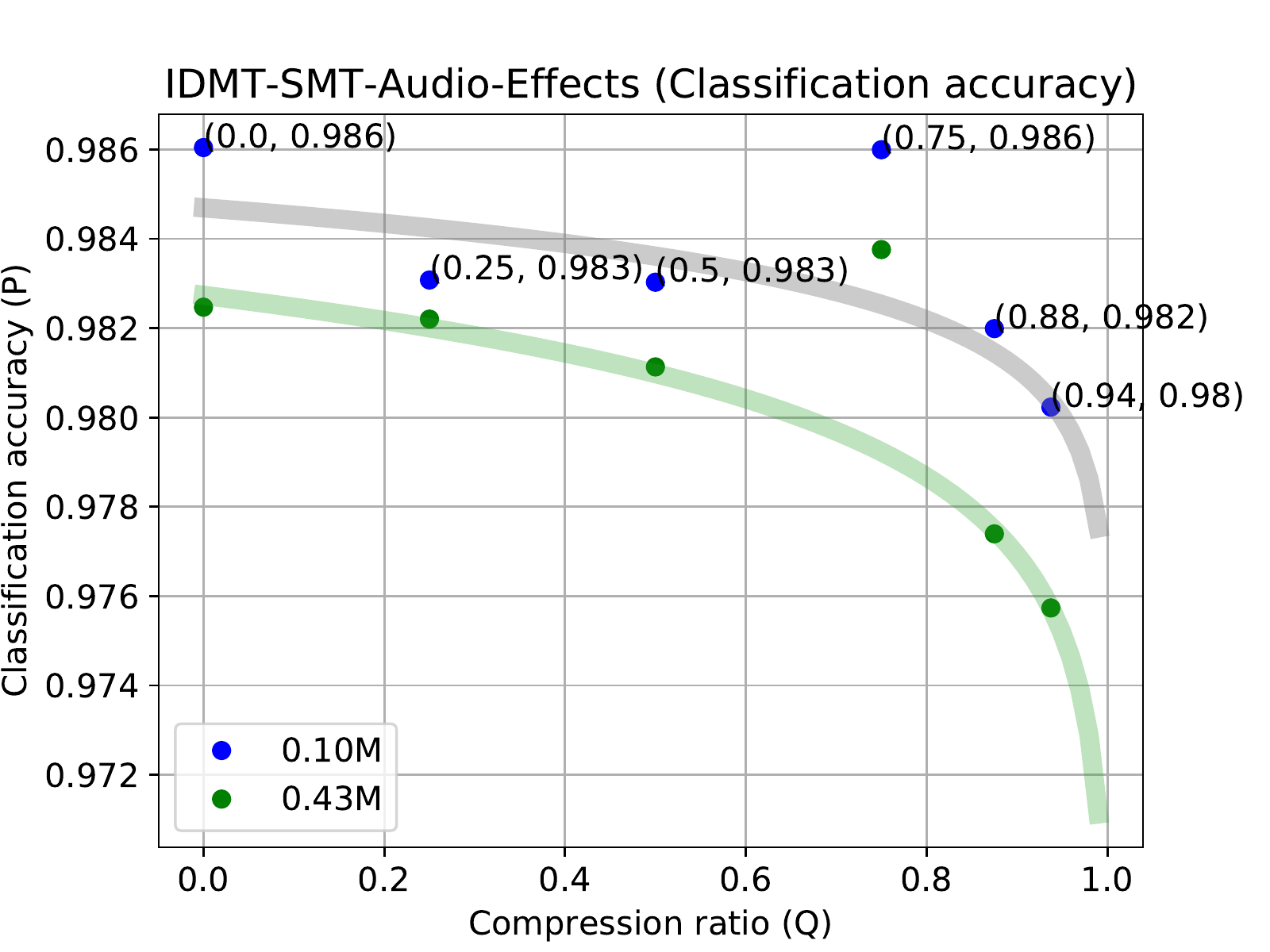}
\caption{\label{fig:audio_acc} Classification accuracy on the audio classification task as a function of the MP3 compression ratio (relative bitrate) on the results of classifier setup A (blue dots) and classifier setup C (green dots). The shadow curves represent the properly scaled version of the theoretical curve from Figure~\protect\ref{fig:qopttheory}. Note that audio compression uses non-linear quantization. Consequently, there is a higher deviation from the hypothesized curve compared to image compression. We normalize the x-axis between the lowest-quality quantization $Q=1.0$ and the highest audio-quality level $Q=0.0$).}
\end{figure}

\begin{figure}
\includegraphics[width=0.48\textwidth]{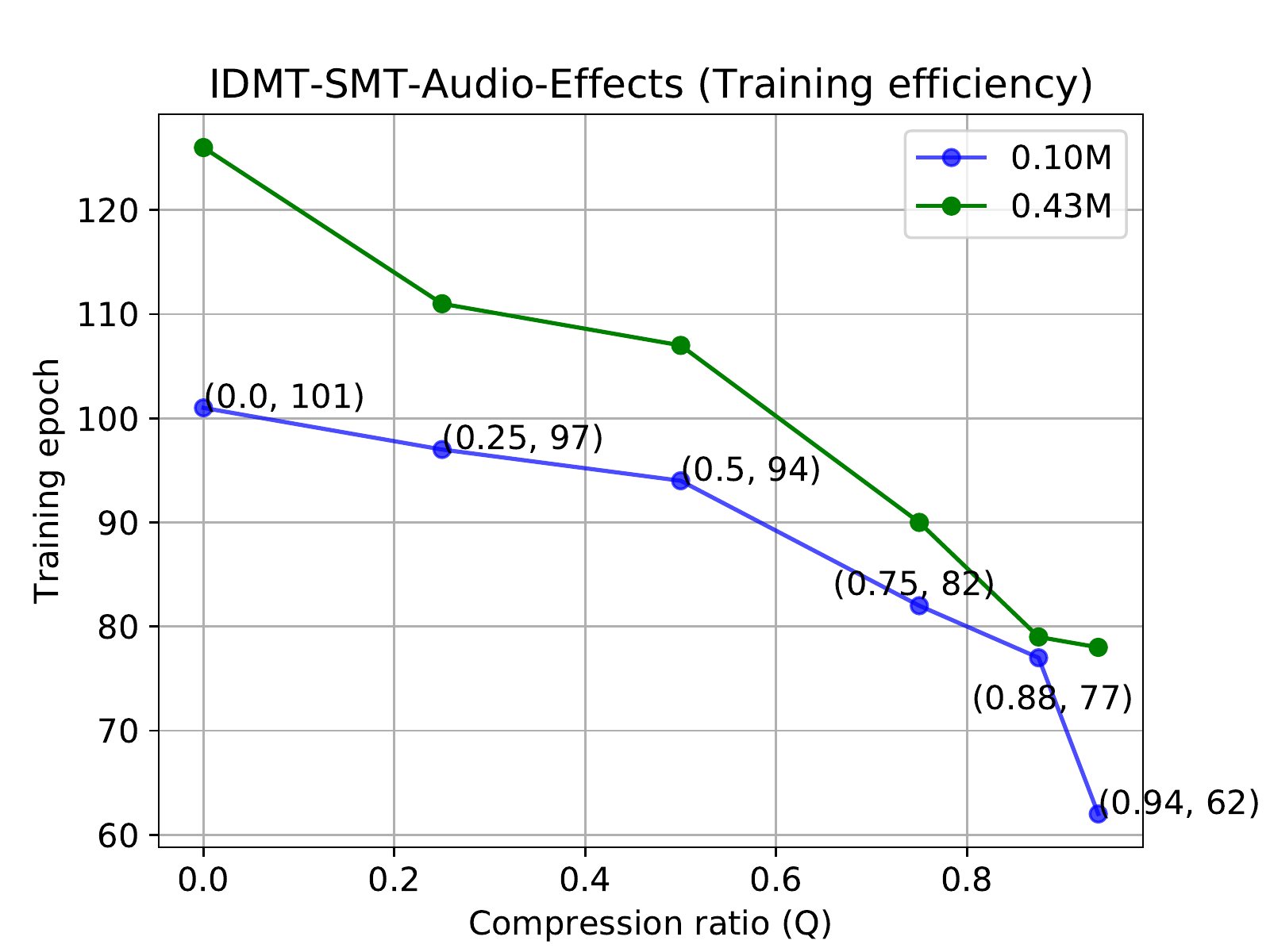}
\caption{\label{fig:audio_time} Number of training epochs on the audio classification task as a function of the MP3 compression ratio (with lowest-quality quantization level being $Q=1.0$).}

\end{figure}

\begin{itemize}
\item A: Conv-loop $\times 3$ and no FC-loop. Size of parameters: 0.10M (\textbf{Blue dots in Figure~\ref{fig:audio_acc} and ~\ref{fig:audio_time}})
\item B: Conv-loop $\times 4$ and FC-loop (128). Size of parameters: 0.17M
\item C: Conv-loop $\times 3$ and FC-loop (64-128). Size of parameters: 0.43M (\textbf{Green dots in Figure~\ref{fig:audio_acc} and ~\ref{fig:audio_time}})
\item D: Conv-loop $\times 3$ and FC-loop (128). Size of parameters: 0.81M
\item E: Conv-loop $\times 3$ and FC-loop (128-128). Size of parameters: 0.82M
\item F: Conv-loop $\times 2$ and FC-loop (128). Size of parameters: 3.72M
\end{itemize}

\begin{table*}
    \centering
	\caption{Neural network architectures used on the audio classification task. Conv: convolutional layer, FC: fully connected layer, BatchNorm: batch normalization layer.}
	\begin{tabular}{@{}ccc@{}}
	\toprule
		\textbf{A} & \textbf{E} & \textbf{F} \\ \hline \hline
		Conv(32, 3, 3) & Conv(32, 3, 3) & Conv(32, 3, 3) \\
		BatchNorm & BatchNorm & BatchNorm \\
		ReLU & ReLU & ReLU \\
		\multirow{4}{*}{$\begin{bmatrix}\mathrm{Conv(32, 3, 3)} \\\mathrm{ELU} \\\mathrm{Maxpooling} \\\mathrm{Dropout(0.5)}\end{bmatrix} \times 3$}
			 & \multirow{4}{*}{$\begin{bmatrix}\mathrm{Conv(32, 3, 3)} \\\mathrm{ELU} \\\mathrm{Maxpooling} \\\mathrm{Dropout(0.5)}\end{bmatrix} \times 3$} & \multirow{4}{*}{$\begin{bmatrix}\mathrm{Conv(32, 3, 3)} \\\mathrm{ELU} \\\mathrm{Maxpooling} \\\mathrm{Dropout(0.5)}\end{bmatrix} \times 2$} \\
		&  &  \\
		&  &  \\
		&  &  \\
		Flatten & Flatten & Flatten \\
		FC(10) & \multirow{2}{*}{$\begin{bmatrix}\mathrm{FC(128)}\\ \mathrm{Dropout(0.6)}\end{bmatrix} \times 2$} & \multirow{2}{*}{\begin{tabular}[c]{@{}c@{}}FC(128)\\ Dropout(0.6)\end{tabular}} \\
		Softmax &  &  \\
		\multicolumn{1}{l}{} & FC(10) & FC(10) \\
		\multicolumn{1}{l}{} & Softmax & Softmax \\ \midrule
		101,412 (0.10M) & 824,868 (0.82M) & 3,715,460 (3.72M) \\ \bottomrule
	\end{tabular}
	\label{tab:audio_arch}
\end{table*}

\subsection{Epoch Measurements}
For both, the image and the audio experiments, Figure~\ref{fig:cifar_time} and~\ref{fig:audio_time} show the number of epochs needed for training as soon as the learner starts to converge. Learning is based on a standard statistic gradient descent for each of the given Deep Neural Network structures. The details of the convergence heuristics can be found on the website mentioned in Section~\ref{sec:conclusion}.

\subsection{Experimental Findings and Discussion}
As can be observed from Figures~\ref{fig:cifar_acc} and~\ref{fig:audio_acc}, both image and audio experiments follow the trend predicted in Section~\ref{sec:perceptual}. That is: 
\begin{figure}
\includegraphics[width=0.48\textwidth]{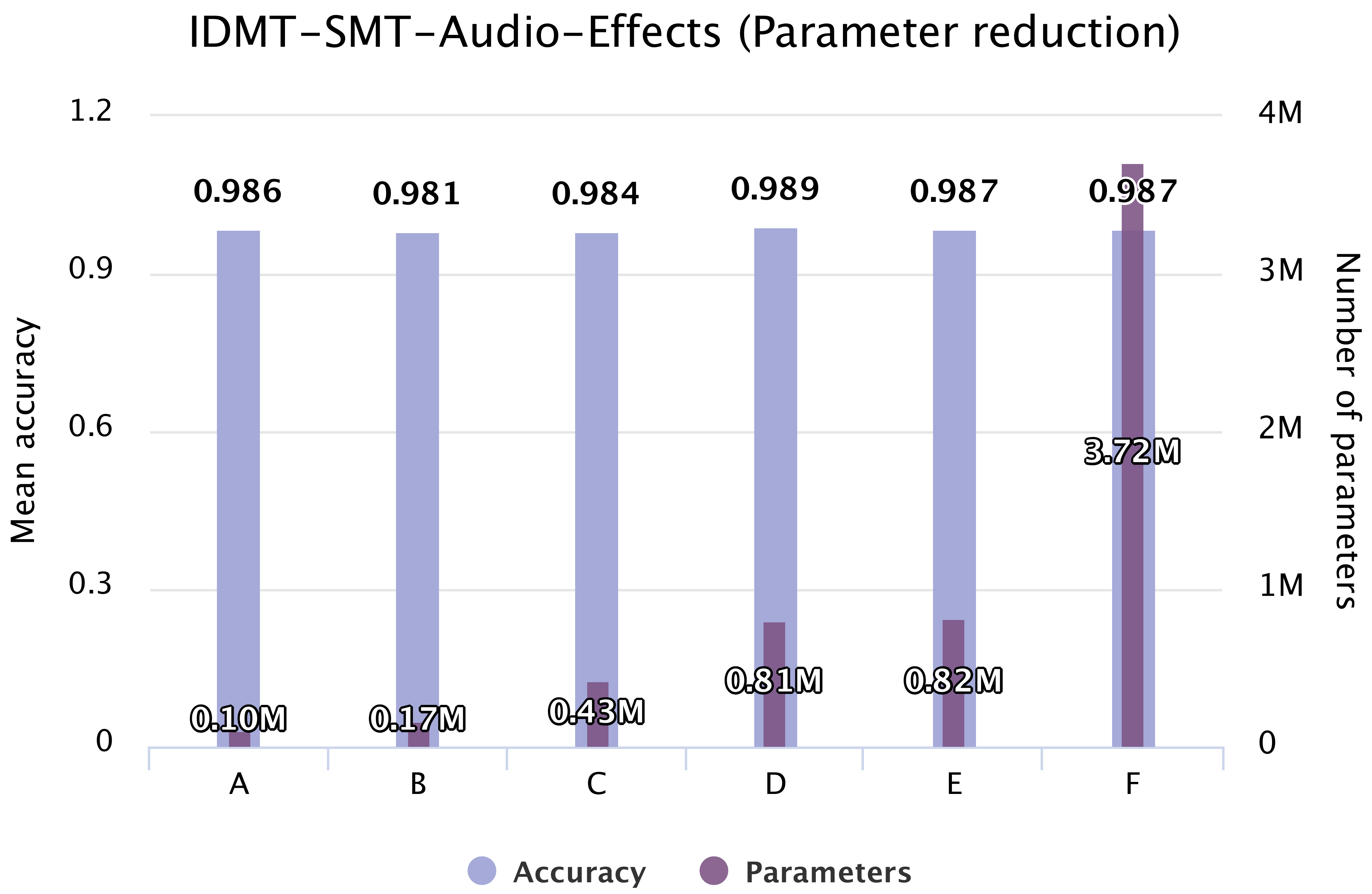}
\caption{\label{fig:audio_paras} Classification accuracy on the audio classification task as a function of the number of parameters using relative quantization level $Q=0.75$ (with lowest-quality quantization level being $Q=1.0$).}
\end{figure}

\begin{enumerate}
    \item The classification accuracy on real data under different compression ratios exactly follows the trend of the hypothesized accuracy curve derived from Equation~\protect\ref{eq:newhelmholtz}. As a consequence, the empirically measured $N_{approx}$ matches the theoretically calculated $N_{approx}$ well for both image and audio datasets.
    \item If the quantization level is smaller than $N_{approx}$, perceptual compression does not seem to affect the classification accuracy significantly.
    \item Due to the reduction of search space implied by a smaller number of parameters, there is a higher chance to find a better set of parameters with the same amount of training. As a consequence, our models with a small number of parameters (blue dots) can even achieve a higher classification accuracy than those with a larger number of parameters (green dots).
\end{enumerate}

As expected, Figure~\ref{fig:cifar_time} and~\ref{fig:audio_time} show that with larger set of parameters, our models needs more training epochs to converge. That is, our empirical results indicate that a quantification of the noise content as outlined in this article is useful to reduce the complexity of machine learning: By controlling the level of perceptual compression, we are able to both achieve high learning utility and reduce training complexity. 

\section{Conclusion and Future Work}
\label{sec:conclusion}
It is easy to treat pixels or audio samples as given and not ask where they come from. However, there is a chain of production that creates the content we are using to investigate machine learning algorithms on multimedia data. In order to understand that chain we need to go back to the fundamentals: Physics ultimately defines what and how we can measure. Applying what is known from physics to our standard methods in machine learning and multimedia computing then allows us to (approximately) measure the signal to noise ratio of any sensor reading which in turn can be used to optimize our machine learning process. 

Our measurements show that only an average of $1.4$ bits per pixel are actually needed to make classification decisions. Passing pixels unfiltered into a deep learning mechanism therefore means that, before the machine learner can recognize patterns of pixels, we need to reduce most of the lesser significant bits before we can get to the significant information per pixel. This is consistent with current attempts to theorize deep learning~\cite{tishby2015deep} and suggest that most of the initial layers of a deep neural network are trained to do noise quantization. Having said that, since noise is relative to a target function, this type of quantization might lead to more accurate results than the generic quantization performed by perceptual compression. Then again, our results indicate that one can potentially save a lot of training time by doing a moderate to aggressive perceptual quantization first.

Future work in this area of multimedia fundamentals could use measurement tools like ours to explore statistics across media boundaries. For example, to quantify the average number of bits needed to distinguish a dog from a cat in images vs in audio vs text. We also expect that measuring the noise content of images can help explain and identify adversarial examples such as described in~\cite{evtimov2017robust}. Overall, we hope that our paper contributes to the fundamentals of our field and encourages other researchers to favor measurements over tuning hyper parameters. 

For the purpose of reproducibility, all our results are publicly available at: \url{https://github.com/wangjksjtu/Helmholtz-DL}

\section*{Acknowledgements}
This work was performed under the auspices of the U.S. Department of Energy by Lawrence Livermore National Laboratory under Contract DE-AC52-07NA27344. It was partially supported by a Lawrence Livermore Laboratory Directed Research \& Development grant (18-ERD-021, ``Explainable AI"). IM release number LLNL-CONF-751806. This work was also supported in part by the Republic of Singapore's National Research Foundation through a grant to the Berkeley Education Alliance for Research in Singapore (BEARS) for the Singapore-Berkeley Building Efficiency and Sustainability in the Tropics (SinBerBEST) Program. BEARS has been established by the University of California, Berkeley as a center for intellectual excellence in research and education in Singapore. Furthermore, this work has also supported by the National Science Foundation under award numbers CNS-1238959, CNS-1238962, CNS-1239054, CNS-1239166, CNS 1514509, and CI-P 1629990. Any findings and conclusions are those of the authors, and do not necessarily reflect the views of the funders. We want to cordially thank Alfredo Metere, Jerome Feldman, Kannan Ramchandran, Bhiksha Raj, and Nathan Mundhenk for their insightful advise. 

\bibliographystyle{abbrv}
\bibliography{helmholtz}

\end{document}